%% file: main.tex
\documentclass[conference]{IEEEtran}
\IEEEoverridecommandlockouts

\usepackage{cite}
\usepackage{amsmath,amssymb,amsfonts}
\usepackage{graphicx}
\usepackage{xcolor}
\usepackage{booktabs}
\usepackage{array}
\newcolumntype{D}[1]{>{\raggedright\arraybackslash}p{#1}}
\usepackage{tikz}
\usetikzlibrary{arrows.meta,positioning,calc,fit,backgrounds}
\usepackage{pgfplots}
\pgfplotsset{compat=1.18}
\usepackage[hidelinks]{hyperref}
\hypersetup{
  pdftitle={FORCA: A Force-Aware Coding Agent for Contact-Rich Robot Manipulation},
  pdfauthor={Xiao Tan, Yixuan Wang, Zhexi Wang, Wenzhen Yuan},
  pdfkeywords={Force and tactile sensing, manipulation planning, embodied agents, contact-rich control}
}

\makeatletter
\def\@IEEEsectpunct{\ \,}
\makeatother

\newcommand{\sys}{\textsc{Forca}}

\begin{document}

\title{FORCA: A Force-Aware Coding Agent for Contact-Rich Robot Manipulation}

\author{\IEEEauthorblockN{Xiao Tan$^{1,*}$, Yixuan Wang$^{1,*}$, Zhexi Wang$^{1,2}$, Wenzhen Yuan$^{1}$}%
\thanks{$^{*}$Xiao Tan and Yixuan Wang contributed equally.}%
\thanks{\raggedright $^{1}$All authors are with the University of Illinois Urbana-Champaign, Urbana, IL, USA.
\texttt{\{tomtan2,yixuan41,zhexiw,yuanwz\}@illinois.edu}.}%
\thanks{\raggedright $^{2}$Zhexi Wang is also with Tsinghua University, Beijing, China.
\texttt{wangzhex23@mails.tsinghua.edu.cn}.}}

\maketitle

\begin{abstract}
Coding agents can translate language instructions into robot programs. For contact-rich manipulation, a key challenge is to generate and refine control logic that responds to physical interaction. We present \sys{}, a coding agent that takes visual observations and wrist force/torque (F/T) measurements as input to write Python programs for contact probing and force-responsive motion. Between executions, the agent inspects force records alongside visual outcomes to diagnose failures and revise both motion and force-feedback control logic. \sys{} requires no demonstrations, task-specific training, or simulation. Across five real-robot tasks, we demonstrate its effectiveness in three forms of contact-rich manipulation: improving manipulation accuracy, adapting to changes in object or surface properties, and regulating contact under surface variation and object motion. Compared with a vision-only variant of the same agent, \sys{} improves grasp accuracy and task success and reduces contact loss in wiping and peeling. 

\end{abstract}


\input{Sections/1-Introduction}

\input{Sections/2-Related_Work}
\input{Sections/3-Method}

\input{Sections/4-Experiments}

\input{Sections/5-Conclusion}

\bibliographystyle{IEEEtran}
\bibliography{refs}

\end{document}

%% file: Sections/1-Introduction.tex
\section{Introduction}
\label{sec:intro}

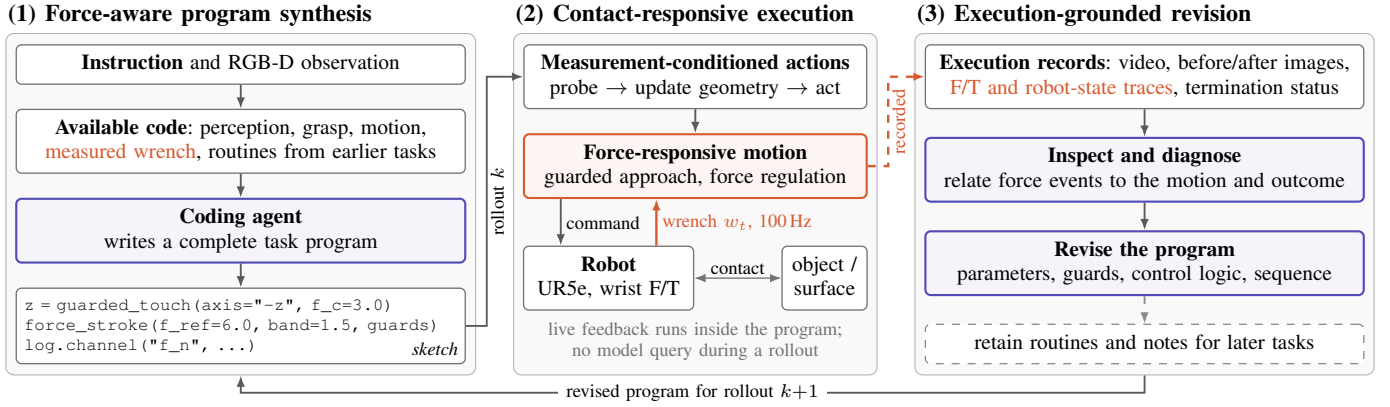
\begin{figure*}[!t]
    \centering
    \input{figures/method_overview}
    \caption{\textbf{Overview of \sys{}.}
    The agent generates a task program that uses force feedback during execution. Recorded force, robot-state, and visual observations inform revisions between rollouts. Generated routines and notes are retained for later reuse.}
    \label{fig:method:overview}
\end{figure*}

Coding agents based on foundation models turn open-ended instructions into executable robot programs that compose perception, grasping, and motion skills \cite{liang2023codeaspolicies,singh2023progprompt,huang2023voxposer,huang2024rekep}. They ground their decisions mainly in vision and proprioception, which is often insufficient for contact-rich manipulation: occluded or reflective surfaces yield uncertain geometry, mass and friction are not visible, and maintaining contact requires responding to surface variation and object motion.

Force and contact measurements complement vision by providing evidence about contact states and mechanical responses that are difficult to infer from visual observations alone. Learned policies exploit force and tactile signals for reactive control but require demonstrations or robot data \cite{foar2024,forcevla2025,tla2025,tactilevla2025}. Foundation-model agents have used force through compliant or force-thresholded primitives \cite{genchip2024,ellmer2025}, inferred physical properties largely from priors \cite{deligrasp2024,llmphy2024,coral2026}, and revised force targets after failed attempts \cite{xie2025unfettered}. Supporting these diverse tasks requires agents to determine when contact information is needed, how it should guide ongoing motion, and how unexpected contact outcomes should inform subsequent attempts.

We present \sys{}, a training-free coding agent that uses wrist force/torque (F/T) sensing in both \emph{program execution} and \emph{program revision} (Fig.~\ref{fig:method:overview}). It writes complete Python programs with access to the measured wrench. During a rollout, the program uses measurements from contact probes to guide subsequent actions and uses live force feedback to regulate its motion. Between rollouts, the agent relates the recorded force traces to visual outcomes, diagnoses failures, and revises the program.

\sys{} addresses three types of contact-rich manipulation: (i)~\emph{precision manipulation}, where force cues help refine grasp poses and improve manipulation accuracy; (ii)~\emph{physical adaptation}, where actions must be adjusted to changes in object or surface properties; and (iii)~\emph{contact regulation}, where motion must respond to surface variation and object movement. We evaluate \sys{} on five real-robot tasks spanning these categories: drawer opening, door-handle operation, cube-to-target knocking, vase wiping, and cucumber peeling. Compared with a vision-only variant of the same agent, \sys{} increases task success from 40\% to 100\% for drawer opening and from 10\% to 100\% for door-handle operation, while reducing contact-loss rates in wiping and peeling from 40.6--55.9\% to at most 2.2\%.

Our contributions are twofold. First, we introduce a coding-agent framework that generates task-specific motion and force-feedback control logic from language instructions. The resulting programs use measured contact to refine grasp poses and regulate motion during execution. Second, we extend program refinement to the force-feedback logic itself, enabling coding agents to revise how their programs respond to contact based on evidence from real-robot execution.

%% file: figures/method_overview.tex
\begin{tikzpicture}[font=\fontsize{8}{9.2}\selectfont, >={Latex[length=1.5mm,width=1.15mm]}, line width=0.5pt]
  \definecolor{force}{HTML}{D85A30}
  \definecolor{acc}{HTML}{534AB7}
  \tikzset{
    panel/.style={draw=black!30, rounded corners=3pt, fill=black!2.5, inner xsep=3.5pt, inner ysep=4pt},
    ptitle/.style={font=\fontsize{9}{10}\selectfont\bfseries, anchor=south west, inner sep=0.8pt},
    box/.style={draw=black!55, fill=white, rounded corners=2pt, align=center, inner xsep=3pt, inner ysep=2.8pt,
                text width=#1, execute at begin node=\strut, execute at end node=\strut},
    agent/.style={box=#1, draw=acc, line width=0.8pt, fill=acc!6},
    fbox/.style={box=#1, draw=force, line width=0.8pt, fill=force!6},
    fl/.style={->, draw=black!65, line width=0.7pt},
    ff/.style={->, draw=force, line width=0.9pt},
    lbl/.style={font=\fontsize{7}{8}\selectfont, inner sep=1.2pt, align=center},
  }
  \def\wA{163pt}\def\wB{123pt}\def\wC{159pt}\def\cgap{13.8pt}
  \def\vgap{9pt}\def\vgapC{10.37pt}
  \pgfmathsetlengthmacro\xa{0.5*\wA+7.15pt}
  \pgfmathsetlengthmacro\xb{\xa+0.5*\wA+0.5*\wB+14.3pt+\cgap}
  \pgfmathsetlengthmacro\xc{\xb+0.5*\wB+0.5*\wC+14.3pt+\cgap}
  \node[box=\wA, anchor=north] (inst) at (\xa,0) {\textbf{Instruction} and RGB-D observation};
  \node[box=\wA, below=\vgap of inst] (lib) {\textbf{Available code}: perception, grasp, motion,\\
                                       \textcolor{force}{measured wrench}, routines from earlier tasks};
  \node[agent=\wA, below=\vgap of lib] (llm) {\textbf{Coding agent}\\ writes a complete task program};
  \node[box=\wA, below=\vgap of llm, font=\ttfamily\fontsize{6.5}{7.6}\selectfont, align=left] (prog)
        {z = guarded\_touch(axis="-z", f\_c=3.0)\\
         force\_stroke(f\_ref=6.0, band=1.5, guards)\\
         log.channel("f\_n", ...)};
  \node[lbl, font=\fontsize{7}{8}\selectfont\itshape, anchor=south east] at ([shift={(-1.5pt,2pt)}]prog.south east) {sketch};
  \draw[fl] (inst)--(lib); \draw[fl] (lib)--(llm); \draw[fl] (llm)--(prog);
  \node[box=\wB, anchor=north] (probe) at (\xb,0) {\textbf{Measurement-conditioned actions}\\ probe $\rightarrow$ update geometry $\rightarrow$ act};
  \node[fbox=\wB, below=\vgap of probe] (reg) {\textbf{Force-responsive motion}\\ guarded approach, force regulation};
  \node[lbl, text=black!55, anchor=south] (note) at (\xb,0 |- prog.south) {live feedback runs inside the program;\\ no model query during a rollout};
  \node[box=58pt, anchor=south west] (robot) at ([yshift=3pt]note.north -| reg.west) {\textbf{Robot}\\ UR5e, wrist F/T};
  \node[box=26pt, anchor=south east] (scene) at ([yshift=3pt]note.north -| reg.east) {object /\\ surface};
  \draw[fl] (probe)--(reg);
  \draw[fl] ([xshift=-18pt]robot.north |- reg.south) -- node[lbl,right=0.6pt]{command} ([xshift=-18pt]robot.north);
  \draw[ff] ([xshift=18pt]robot.north) -- node[lbl,right=0.6pt,text=force]{wrench $w_t$, 100\,Hz} ([xshift=18pt]robot.north |- reg.south);
  \draw[<->, draw=black!55] (robot) -- node[lbl,above]{contact} (scene);
  \node[box=\wC, anchor=north] (rec) at (\xc,0) {\textbf{Execution records}: video, before/after images,\\
                                       \textcolor{force}{F/T and robot-state traces}, termination status};
  \node[agent=\wC, below=\vgapC of rec] (diag) {\textbf{Inspect and diagnose}\\ relate force events to the motion and outcome};
  \node[agent=\wC, below=\vgapC of diag] (rev) {\textbf{Revise the program}\\ parameters, guards, control logic, sequence};
  \node[box=\wC, dashed, draw=black!45, anchor=south] (keep) at (\xc,0 |- prog.south) {retain routines and notes for later tasks};
  \draw[fl] (rec)--(diag); \draw[fl] (diag)--(rev); \draw[fl, dashed, draw=black!45] (rev)--(keep);
  \begin{scope}[on background layer]
    \node[panel, fit=(inst)(lib)(llm)(prog)] (P1) {};
    \node[panel, fit=(probe)(reg)(robot)(scene)(note)] (P2) {};
    \node[panel, fit=(rec)(diag)(rev)(keep)] (P3) {};
  \end{scope}
  \node[ptitle] at ([yshift=1.5pt]P1.north west) {(1) Force-aware program synthesis};
  \node[ptitle] at ([yshift=1.5pt]P2.north west) {(2) Contact-responsive execution};
  \node[ptitle] at ([yshift=1.5pt]P3.north west) {(3) Execution-grounded revision};
  \draw[fl] (prog.east) -- ([xshift=4pt]prog.east -| P1.east) |- node[lbl,pos=0.25,rotate=90,anchor=north]{rollout $k$} (probe.west);
  \draw[ff, dashed] (reg.east) -- ([xshift=4pt]reg.east -| P2.east) |- node[lbl,pos=0.25,rotate=90,anchor=north,text=force]{recorded} (rec.west);
  \draw[fl] (P3.south) -- ++(0,-6.5pt) coordinate (q) -- (q -| P1.south)
            node[lbl,pos=0.5,fill=white,inner xsep=3pt,inner ysep=1pt]{revised program for rollout $k{+}1$} -- (P1.south);
\end{tikzpicture}

%% file: Sections/2-Related_Work.tex
\section{Related Work}
\label{sec:related}

\noindent\textbf{Foundation-model agents for manipulation.} Foundation models generate robot programs and plans from language \cite{liang2023codeaspolicies,singh2023progprompt}, ground manipulation in value maps or keypoint constraints \cite{huang2023voxposer,huang2024rekep}, and revise plans from scene descriptions, success detectors, or language feedback \cite{huang2023innermonologue}; recent work benchmarks coding agents for manipulation \cite{fu2026capx} and refines programs or policies from real-robot experience \cite{aspire2026,enpire2026}. Their feedback is visual, proprioceptive, or linguistic. Agents that do use force expose compliant, force-thresholded primitives to code generation \cite{genchip2024}, call force-feedback routines from a knowledge base \cite{ellmer2025}, or have a VLM output target wrenches that it revises after a failed attempt \cite{xie2025unfettered}. To our knowledge, \sys{} is the first coding-agent framework to jointly generate task-specific force-feedback control logic and revise that logic between real-robot rollouts using recorded F/T traces and visual outcomes.

\noindent\textbf{Contact-based perception and physical adaptation.} Contact complements vision with local observations of geometry and interaction state: visuotactile estimators fuse vision, proprioception, and force under occlusion \cite{lee2020multimodal}, and active tactile exploration selects informative contacts to estimate pose and shape \cite{gordon2026active}. Interaction also reveals latent properties, estimated by active force-based methods \cite{activemass2023} or inferred across repeated trials by a learned memory \cite{gao2026gmp}; these methods address a fixed estimation problem with exploration designed or learned for it. Foundation-model approaches infer properties largely from priors \cite{deligrasp2024,xie2025unfettered}, by matching observations in a physics engine \cite{llmphy2024}, or through online system identification \cite{coral2026}, and Matcha \cite{zhao2023matcha} lets an LLM choose exploratory actions for material recognition in simulation. In \sys{}, the agent decides at deployment time when a visual estimate is unreliable and where to touch, and adapts a developed program to changed physical conditions from measured interactions, without task-specific training.

\noindent\textbf{Force feedback for reactive manipulation.} Force feedback lets robots adjust their motion as contact conditions change. It is widely used in compliant control and guarded motion \cite{hogan1985impedance,whitney1982peg,will1975assembly}. Learned policies add force or tactile feedback to visuomotor control \cite{foar2024,xue2025reactive,fang2026forcepolicy,xu2026contactgrounded} or fuse it into vision-language-action models \cite{forcevla2025,tla2025,tactilevla2025}. Classical methods rely on thresholds and nominal traces specified per task, and learned policies require demonstrations or robot data. \sys{} builds on the same control principles but has the agent write the force-responsive controller at deployment time and revise it from the recorded F/T trace.

%% file: Sections/3-Method.tex
\section{Method}
\label{sec:method}

Given a language instruction and RGB-D observations, \sys{} writes a Python code with access to wrist force/torque (F/T) measurements. The program executes on the robot without querying the foundation model. After execution, the agent reviews the recorded interaction and revises the code (Fig.~\ref{fig:method:overview}). We refer to one execution of a program as a \emph{rollout}.

\subsection{Force-Aware Program Synthesis}
\label{sec:method:synthesis}

\begin{figure*}[!t]
    \centering
    \includegraphics[width=\textwidth]{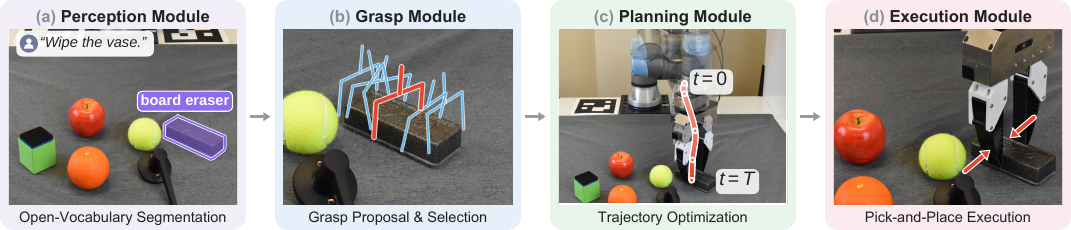}
    \caption{\textbf{Predefined skill modules.}
    The agent is initialized with these modules, assembled from open-source components under human supervision.
    (a)~SAM~3~\cite{carion2026sam3} localizes the tool named in the instruction.
    (b)~GraspGen~\cite{graspgen2025} proposes candidate grasps, and the VLM selects one.
    (c)~PyRoKi~\cite{pyroki2025} plans the trajectory.
    (d)~The robot executes the resulting routine to acquire the tool.}
    \label{fig:method:skills}
\end{figure*}

For each task, the agent generates a Python task program that specifies the action sequence and the force-dependent motion logic. To construct this program, the agent has access to an initial skill codebase for perception, grasping, motion planning, and pick-and-place, assembled from open-source components under human supervision. These predefined modules provide object pose estimates, grasp candidates, and collision-free trajectories computed with PyRoKi~\cite{pyroki2025}. Figure~\ref{fig:method:skills} illustrates these modules as the robot acquires a whiteboard eraser for the instruction ``Wipe the vase.''

The agent writes the force-dependent logic for each task using wrist wrench measurements exposed by the robot interface. For example, this logic can specify when an approach should stop or how motion should respond to a force error. To implement these behaviors, the agent can modify existing routines or add measurement and control code as needed.

\begin{figure*}[!t]
    \centering
    \includegraphics[width=\textwidth]{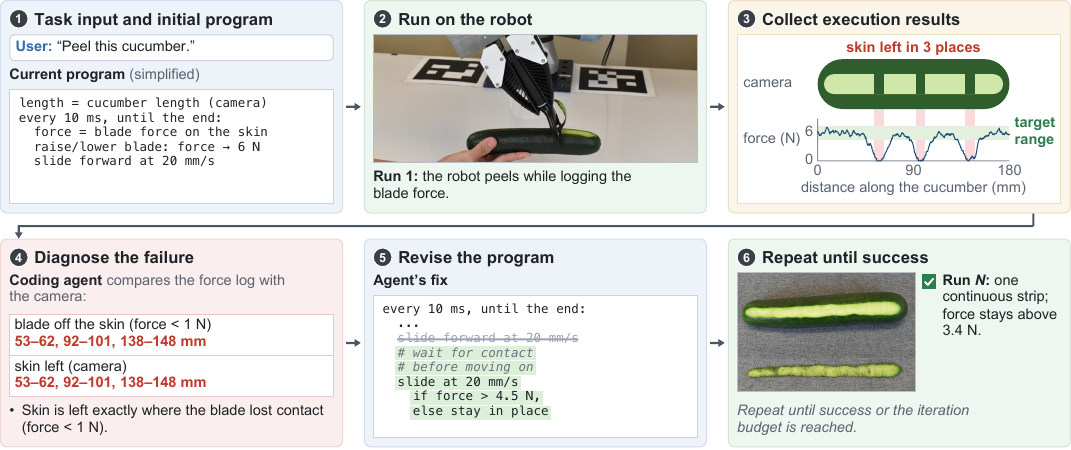}
    \caption{\textbf{Illustrative program revision for cucumber peeling.}
    The agent compares force records with visible peeling gaps and adds a condition that pauses forward motion until contact is restored.
    Photos are from real runs; force traces and skin patches are idealized, and the code is simplified.}
    \label{fig:method:refinement}
\end{figure*}

\subsection{Contact-Responsive Execution}
\label{sec:method:execution}
During a rollout, the generated program uses F/T measurements to guide subsequent actions and adjust ongoing motion. The conditions and feedback rules governing these responses are encoded in the program and executed without querying the foundation model.

Contact measurements can inform actions by refining the geometric estimates on which they depend. For example, the program can probe a surface before grasping, specifying both the probing motion and the force condition used to detect contact. The robot pose at contact, together with the tool geometry, provides an updated estimate of surface height or grasp position. Subsequent motions use this estimate within the same rollout.

For actions that require sustained contact, the program uses live force feedback to regulate motion. A host-side 100\,Hz loop reads the wrench and robot state over RTDE and updates Cartesian velocity commands according to the generated control logic. This logic can adjust motion based on the difference between measured and target force, or pause forward motion when contact is insufficient and resume it once contact is restored.

\subsection{Execution-Grounded Program Revision}
\label{sec:method:revision}
After each rollout, the agent inspects visual observations alongside the recorded force and robot-state traces. The agent can write analysis code to examine relevant time windows and relate force changes to robot motion and visible outcomes.

The agent uses these records to identify possible causes of failure and revise the relevant code. Changes can affect motion parameters or contact conditions, and can also modify the control logic or action sequence. 
Figure~\ref{fig:method:refinement} illustrates one such revision in cucumber peeling. Skin remains exactly where the recorded force shows that the blade lost contact, so the agent makes the blade wait for contact before moving on.

Before execution, checks screen the revised code and planned motions for errors, reachability, and workspace violations. Revision stops when the agent declares the program converged or when either the rollout or wall-clock budget is exhausted. Development budgets and evaluation protocols are described in Sec.~\ref{sec:exp:setup}.

%% file: Sections/4-Experiments.tex
\section{Experiments}
\label{sec:experiments}

\begin{figure*}[!t]
    \centering
    \includegraphics[width=\textwidth]{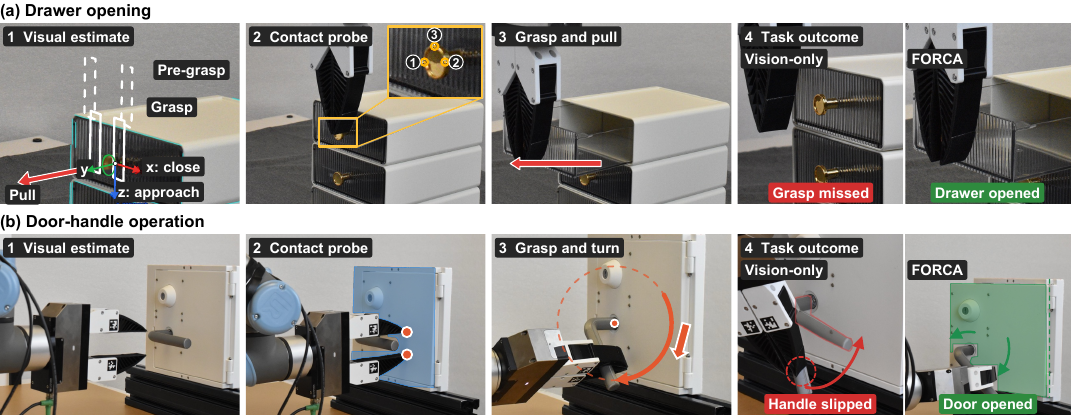}
    \caption{\textbf{Example rollouts and outcome comparisons.}
    (a)~Drawer opening and (b)~door-handle operation.
    Panels~1--3 show \sys{} estimating geometry visually, refining it through contact probing, and grasping to pull the drawer or turn the handle.
    In~(a1), dashed and solid outlines mark the estimated pre-grasp and grasp poses; the axes indicate closing and approach directions.
    Numbered markers in~(a2) identify three probe locations on the drawer knob.
    In~(b2), orange markers indicate fingertip contacts, and blue shading marks the door-panel position confirmed by contact.
    Panels~(a4) and~(b4) compare example outcomes: the vision-only agent misses the drawer grasp and loses its grip on the door handle, while \sys{} opens both mechanisms.
    The dashed circle and red arrow in~(b4) mark the fingertip-only grasp and the handle's rebound after slipping.}
    \label{fig:exp:manipulation}
\end{figure*}

We consider three ways in which force/torque feedback can benefit manipulation: (i)~\emph{improving manipulation accuracy}, by complementing visual estimates with contact-based geometric information to refine grasp poses and guide subsequent actions; (ii)~\emph{adapting to physical changes}, by providing evidence about how object properties affect action outcomes to adjust manipulation accordingly; and (iii)~\emph{regulating contact}, by using live force measurements to adjust ongoing motion and maintain contact as surfaces vary or objects move. Our experiments evaluate how effectively \sys{} uses contact information across these three settings, combining comparisons against a vision-only agent with case studies of the generated programs. We additionally examine how \sys{} revises its programs across iterations and the time and robot interactions required.

\subsection{Experimental Setup and Protocol}
\label{sec:exp:setup}

\noindent\textbf{Hardware and agent.} Experiments use a UR5e arm with wrist force/torque sensing, a WSG-50 gripper, and a fixed RealSense D455 RGB-D camera. The coding agent uses Claude Fable 5 for all analysis and code writing. The initial skills are described in Sec.~\ref{sec:method}.

\noindent\textbf{Execution and evaluation.} The agent generates a complete task program, executes it, inspects the resulting records, and revises the program. One rollout consists of the agent writing or revising the task program and executing it once on the real robot. Each run is limited to 20 rollouts and 4 hours of wall-clock time. The run stops when the agent declares the program converged, after which we freeze the final program and run it over repeated trials and assess the physical task outcome; otherwise, it ends when either limit is reached. A rollout whose execution reaches the hard-coded 30\,N force limit is stopped and counted as a failure.
Human involvement is limited to execution authorization, scene operations, and final acceptance assessment.

\noindent\textbf{Baseline policy.}
We compare \sys{} with a \textbf{vision-only agent} that uses the same foundation model, task instructions, and basic skills, but has no access to force/torque data in program generation, revision, or execution: it receives no live or recorded wrench and no force-responsive control routines. Its programs guide motion with visual observations and robot state, and the agent can write visual feedback code and inspect visual outcomes when revising them. Robot-level safety protections remain active but do not expose force data to the agent, and the force recordings used for offline evaluation are withheld from it.

\subsection{Results and Analysis}
\label{sec:exp:results}

\subsubsection{Tasks Requiring Precise Manipulation}
\label{sec:exp:percept}

We evaluate whether force cues improve manipulation accuracy and task success on \textbf{drawer opening} and \textbf{door-handle operation}. Both tasks require the robot to locate and grasp a handle before actuating the mechanism. The drawer is transparent, making its handle difficult to localize precisely from vision alone. Contact probing provides additional geometric information to refine the grasp pose. We compare \sys{} with the vision-only agent using translational and rotational \emph{grasp error} and \emph{task success rate} (Table~\ref{tab:e1}). For each task and method, we execute the frozen program for 10 trials and report the success rate and grasp errors across these trials. \sys{} achieves lower translational and rotational grasp errors and succeeds in all 10 trials on both tasks (100\%), whereas the vision-only agent succeeds in only 4/10 trials on drawer opening and 1/10 on door-handle operation. Section~\ref{sec:exp:revision} examines how the door-handle programs were developed, with detailed failures and revisions documented in Table~\ref{tab:door}.

\begin{table}[!t]
    \caption{Grasp accuracy and task success for drawer opening and door-handle operation. Grasp errors: mean $\pm$ std.\ over 10 trials.}
    \label{tab:e1}
    \centering
    \footnotesize
    \setlength{\tabcolsep}{4.5pt}
    \begin{tabular}{@{}llccc@{}}
        \toprule
        Task & Method
            & \shortstack{Trans.\ err.\\(mm) $\downarrow$}
            & \shortstack{Rot.\ err.\\($^\circ$) $\downarrow$}
            & \shortstack{Success\\(\%) $\uparrow$} \\
        \midrule
        Drawer & Vision-only & $8.3 \pm 5.4$ & $9.6 \pm 3.2$ & 40 \\
               & \sys{}      & $5.3 \pm 1.6$ & $1.0 \pm 0.3$ & 100 \\
        \midrule
        Door   & Vision-only & $65.3 \pm 16.9$ & $55.5 \pm 14.9$ & 10 \\
               & \sys{}      & $20.0 \pm 0.9$  & $0.6 \pm 0.5$ & 100 \\
        \bottomrule
    \end{tabular}
\end{table}

\subsubsection{Tasks Requiring Adaptation to New Physical Conditions}
\label{sec:exp:physics}

We evaluate physical adaptation on a \textbf{cube-to-target} task, in which the robot knocks an object so as to make it slide to a target region. The robot must adjust its knocking motion to send the object far enough to reach the target without overshooting. The object's mass and contact geometry affect its response to the impact, while friction with the supporting surface influences its subsequent deceleration and stopping distance. Consequently, the same knocking motion can produce different final positions when the object or surface changes.

\begin{figure}[!t]
    \centering
    \includegraphics[width=\columnwidth]{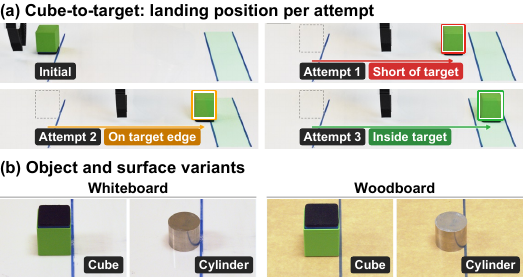}
    \caption{\textbf{Cube-to-target adaptation.}
    (a)~Object positions across successive attempts.
    The dashed outline marks the initial pose, and shading indicates the target region.
    (b)~Object and surface configurations.}
    \label{fig:exp:adaptation}
\end{figure}

We ask whether a program developed under one physical condition can be reused when the object's response changes, and how many additional interactions it needs to complete the task again. We vary the object, using a cube and a cylinder that differ in both shape and mass (61.9\,g and 663.6\,g), and the surface, using a whiteboard and a woodboard that differ in friction (Fig.~\ref{fig:exp:adaptation}(b)). These variations change how far the object travels in response to a knock while preserving the task objective.

\begin{figure}[!t]
    \centering
    \input{figures/exp_adaptation_curve}
    \vspace{-6pt}
    \caption{\textbf{Reuse across physical conditions.} Distance to the target band per attempt (0: inside), starting from the program developed for the cube on the whiteboard.}
    \label{fig:exp:adaptation_curve}
\end{figure}
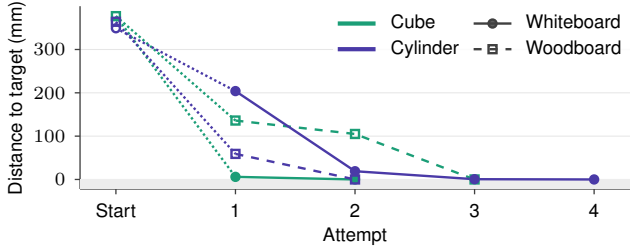

The evaluation has two stages. In the \emph{development} stage, \sys{} builds a knocking program from scratch for the cube on the whiteboard, the reference configuration (Table~\ref{tab:rollouts}). In the \emph{reuse} stage, we change the object or the surface and let \sys{} continue from the developed program. After each attempt, it reviews the measured interaction forces together with the observed final object position and revises the program. Figure~\ref{fig:exp:adaptation_curve} shows the distance from the object to the target band after each attempt, and Fig.~\ref{fig:exp:adaptation}(a) shows illustrative landing positions across attempts. In each of the three new configurations, the retained program reached the target band within a few additional iterations, far fewer than developing it from scratch required (Table~\ref{tab:rollouts}). The interaction experience of \sys{} is thus kept in executable code, which remains usable, and can be revised, when physical conditions change.

\begin{figure*}[!t]
    \centering
    \includegraphics[width=\textwidth]{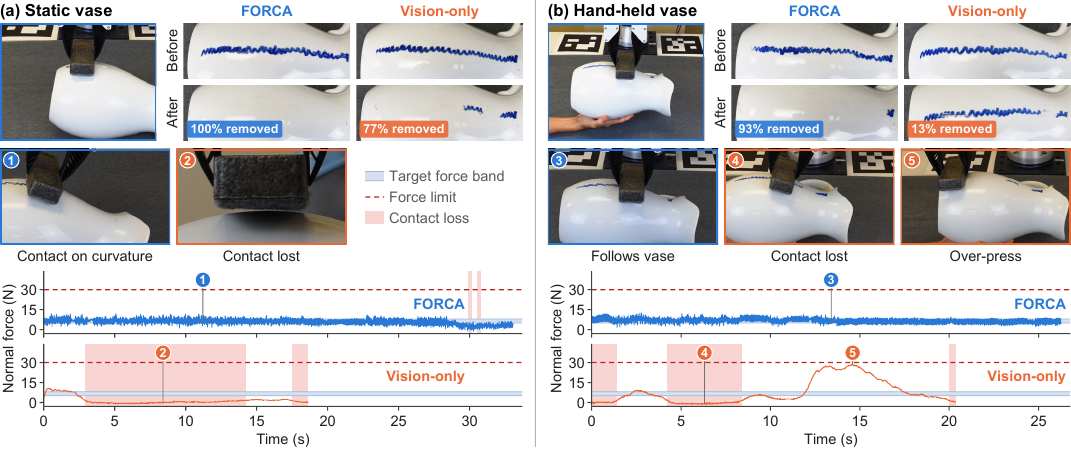}
    \caption{\textbf{Vase wiping: (a)~static vase and (b)~zero-shot transfer to a hand-held vase.} Each panel shows representative outcomes (top), execution frames (middle), and force traces (bottom).
    Both programs are refined in the static setting and transferred without further revision.
    As the vase moves, \sys{} maintains more consistent contact and high marker removal, whereas the vision-only program exhibits contact loss and large force variations, leaving substantial residual marks.}
    \label{fig:exp:wiping}
\end{figure*}

\subsubsection{Tasks Requiring Online Contact Regulation}
\label{sec:exp:reactive}

We evaluate force-responsive execution on \textbf{vase wiping} and \textbf{cucumber peeling}. Wiping tests whether programs refined on a stationary surface remain effective when the object moves. Peeling tests whether the robot can maintain contact along an uneven surface while avoiding excessive force.

\noindent\textbf{Vase wiping.} The robot uses a whiteboard eraser to remove a marker line from a vase. We consider two settings: a \emph{static} vase resting on a table and a \emph{hand-held} vase that a person translates and rotates during wiping. We evaluate both methods after iterative program refinement in the static setting, then deploy the statically refined programs in the hand-held setting without further program revision.

Table~\ref{tab:e3} reports the \emph{marker-removal ratio}, \emph{contact-loss rate}, and \emph{maximum force} for both methods in each setting. The marker-removal ratio is the percentage of the original mark removed, measured from before/after images. After iterative refinement, both methods achieve high marker removal on the static setup. The vision-only program improves by repeatedly adjusting trajectory offsets to the fixed surface, while \sys{} uses force feedback to correct motion during contact. When transferred to the hand-held setting, the tuned offsets become inadequate as the vase moves, reducing removal and increasing contact loss. \sys{} adjusts its motion to the moving surface and maintains more consistent contact. Figure~\ref{fig:exp:wiping} shows the wiping outcomes and corresponding force traces.

\begin{table}[!t]
\caption{Vase wiping results averaged over 10 trials.}
    \label{tab:e3}
    \centering
    \footnotesize
    \setlength{\tabcolsep}{5pt}
    \begin{tabular}{@{}llccc@{}}
        \toprule
        Vase & Method
            & \shortstack{Removal\\(\%) $\uparrow$}
            & \shortstack{Contact\\loss (\%) $\downarrow$}
            & \shortstack{Max.\\force (N) $\downarrow$} \\
        \midrule
        Static    & Vision-only & 73 & 42.6 & 11.8 \\
                  & \sys{}      & 99 & 0.7  & 11.3 \\
        \midrule
        Hand-held & Vision-only & 21 & 55.9 & 17.5 \\
                  & \sys{}      & 93 & 1.6  & 12.3 \\
        \bottomrule
    \end{tabular}
\end{table}

\begin{table}[!t]
    \caption{Peeling a hand-held cucumber (averaged over 10 trials)}
    \label{tab:e3:peeling}
    \centering
    \footnotesize
    \setlength{\tabcolsep}{6pt}
    \begin{tabular}{@{}lccc@{}}
        \toprule
        Method
            & \shortstack{Contact\\loss (\%) $\downarrow$}
            & \shortstack{Excessive\\force (\%) $\downarrow$}
            & \shortstack{Peeling\\ratio (\%) $\uparrow$} \\
        \midrule
        Vision-only & 40.6 & 13.4 & 37.4 \\
        \sys{}      & 2.2  & 2.7  & 81.1 \\
        \bottomrule
    \end{tabular}
\end{table}

\noindent\textbf{Cucumber peeling.} The robot peels a longitudinal strip from a cucumber held by a person. We compare the vision-only agent and \sys{} using \emph{contact-loss rate}, \emph{excessive-force rate}, and \emph{peeling ratio} (Table~\ref{tab:e3:peeling}). The peeling ratio is the total newly peeled length divided by the intended strip length; separated peeled segments contribute their individual lengths, excluding unpeeled gaps. Excessive-force rate measures the proportion of active peeling time above an upper contact-force threshold, capturing sustained over-pressing that a peak-force value alone cannot characterize.

\begin{figure}[!t]
    \centering
    \includegraphics[width=\columnwidth]{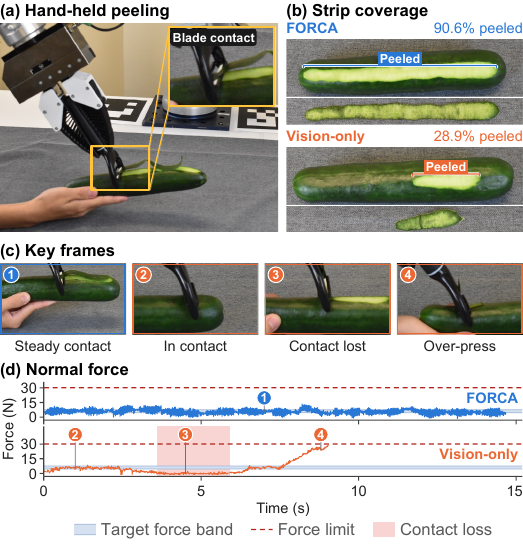}
    \caption{\textbf{Cucumber peeling rollouts.} (a)~Hand-held setup and a magnified view of the boxed blade--surface contact. (b)~Peeled strips after one representative pass, covering 90.6\% of the intended length for \sys{} and 28.9\% for the vision-only agent. (c)~Execution frames at the marked instants. (d)~Normal contact force. Shaded intervals indicate contact loss; the blue band marks the 4.5--7.5\,N target range, and the dashed line marks the 30\,N force limit.}
    \label{fig:exp:peeling}
\end{figure}

\sys{} follows the cucumber surface with force-responsive motion and produces a more continuous peeled strip. The vision-only program exhibits contact interruptions and excessive pressing, leaving gaps in the peeled strip. Figure~\ref{fig:exp:peeling}(a,~b) shows representative peeling outcomes, and Fig.~\ref{fig:exp:peeling}(c,~d) relates the observed contact behavior to the force traces.

\noindent\textbf{Contact metrics.} Pressing force is the tool-normal force projection for wiping and the base-frame $z$ force component for peeling. Contact-loss rate is the percentage of active task time spent in intervals where the pressing force stays below a contact threshold, 1.5\,N for wiping and 0.8\,N for peeling, for at least 0.2\,s. Excessive-force rate uses $F_{\mathrm{high}} = 10$\,N, the upper threshold for acceptable peeling force, and maximum force is the largest pressing-force sample during wiping. Both rates include pauses for contact recovery and exclude the initial approach and deliberate lift-off. Force traces are recorded for offline evaluation with the same definitions for both methods; neither the vision-only agent nor its programs can access them.

\subsection{Program Development Cost} 

Table~\ref{tab:rollouts} reports the number of iterations \sys{} used on each task and the mean time per iteration spent on agent reasoning and on robot motion. Agent reasoning takes most of each iteration, 4--10\,min against 1--3\,min of robot motion. Wiping and peeling used the full budget of 20 iterations.

\begin{table}[!t]
    \caption{Development iterations and mean time per iteration.
    For new cube configurations, the count is the mean number of revisions after the first execution, averaged over three configurations.}
    \label{tab:rollouts}
    \centering
    \footnotesize
    \setlength{\tabcolsep}{3pt}
    \begin{tabular}{@{}lccc@{}}
        \toprule
        Task & Iterations
            & \shortstack{Agent reasoning\\(min / iteration)}
            & \shortstack{Robot motion\\(min / iteration)} \\
        \midrule
        Door handle                          & 17 & 10 & 3   \\
        Drawer                               & 8  & 5  & 1.5 \\
        Cube-to-target, from scratch         & 13 & 4  & 1   \\
        Cube-to-target, new condition        & 2  & 4  & 1   \\
        Vase wiping                          & 20 & 4  & 1   \\
        Cucumber peeling                     & 20 & 5  & 1.5 \\
        \bottomrule
    \end{tabular}
\end{table}

\begin{table*}[!t]
    \caption{Door-handle iterations of \sys{} and the vision-only agent.}
    \label{tab:door}
    \centering
    \scriptsize
    \setlength{\tabcolsep}{3pt}
    \renewcommand{\arraystretch}{1.15}
    \begin{tabular}{@{}D{17.5pt}D{33pt}D{268pt}D{\dimexpr\textwidth-336.5pt\relax}@{}}
        \toprule
        It. & Failed at & What happened & Revision \\
        \midrule
        \multicolumn{4}{@{}l}{\textbf{\sys{}}} \\
        1--2 & Approach
            & Camera put the door 3--4\,cm too close; the probe missed it, then rejected a real touch.
            & Search farther; trust the touch and update the door depth. \\
        3 & Press down
            & First firm grasp, but at the 10\,N force cap the handle turned only 18$^\circ$, too little to unlatch.
            & Press along the handle's arc about its axis. \\
        4--5 & Press down
            & Its own motion checks read slightly short moves as blocked; the handle never passed 5$^\circ$.
            & Detect blocking by force; use the measured handle pose. \\
        6--8 & Press down
            & Still latched at 72.5$^\circ$: grasp 21\,mm and axis 15\,mm off; handle twisted in the fingers.
            & Model the handle; press about a measured rotation axis. \\
        9--10 & Approach, press down
            & Setup faults: the cable plug, missing from the collision model, hit the door frame, and overly strict safety checks stopped motions; the press stopped at 49.7$^\circ$.
            & Fix the setup; in the recorded press force, find latch release (61--65$^\circ$) and end stop (jump past 74.5$^\circ$); press to the stop. \\
        11 & Pull open
            & Latch released at 79.9$^\circ$, but end-stop push-back used up the pull force budget.
            & Ease off one step from the end stop before pulling. \\
        12--13 & Release
            & Door opened (${\sim}26^\circ$); the handle-return move pushed on it; the retreat crossed the door.
            & Open the gripper and back straight out; replan the retreat. \\
        14 & Pull open
            & End stop not detected; push-back again used up the pull force budget.
            & Set the pull force limit relative to the push-back. \\
        15 & Release
            & Door opened and handle released; a leftover force reading blocked the retreat.
            & Re-zero the force sensor after letting go. \\
        16 & Press down
            & In one run, 49.7$^\circ$ was mistaken for the end stop; the robot pulled while still latched.
            & Accept end stops only past 65$^\circ$; pull after the force drops. \\
        17--20 & Success
            & Door opened, handle released, and robot backed away in every rollout.
            & --- \\
        \midrule
        \multicolumn{4}{@{}l}{\textbf{Vision-only}} \\
        1--2 & Approach
            & A self-imposed ${\geq}5$\,cm handle-standoff check failed at 4.9\,cm; the robot never moved.
            & Lower the threshold. \\
        3--6 & Grasp
            & Bar landed at the fingertips (depth ${\sim}1$\,cm off, tilted gripper); it slipped at the first press.
            & Grasp deeper with less tilt; rebuild the handle model. \\
        7--12 & Press down
            & Fingers flexed, so a commanded 35$^\circ$ turned the handle only ${\sim}25^\circ$; the door stayed latched and the bar slipped out when pulled. In 9--10, the bar again landed at the fingertips.
            & Hook one finger behind the bar; raise the commanded press angle from 35$^\circ$ to 70$^\circ$. \\
        13--20 & Pull open
            & Handle reached its end stop (${\sim}62^\circ$), but the bar slipped out of the grip during the pull.
            & Reorient and roll the gripper to hook the bar. \\
        \bottomrule
    \end{tabular}
\end{table*}

\subsection{Iterative Program Revision}
\label{sec:exp:revision}

During iterative revision, the agent uses execution feedback to diagnose failures and update its program between rollouts. To illustrate this process, we present door-handle operation as a case study and analyze one development run of each agent.

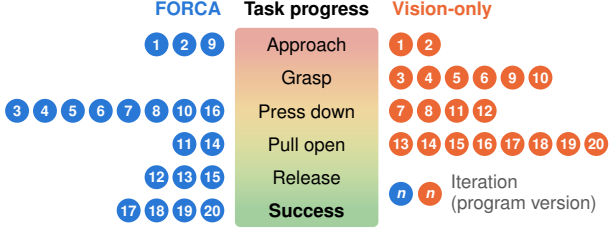
\begin{figure}[!t]
    \centering
    \input{figures/exp_door_iterations}
    \caption{\textbf{Door-handle development histories.}
    Each marker denotes an execution at its failure stage or at task completion.
    Executions~18--20 repeat \sys{}'s final program. Detailed failures and revisions are listed in Table~\ref{tab:door}.}
    \label{fig:exp:door_iterations}
\end{figure}

\noindent\textbf{Door-handle case study.} Figure~\ref{fig:exp:door_iterations} shows representative histories of iterative program development for \sys{} and the vision-only agent. Each marker sits at the stage where that iteration's execution failed, so reading the markers in order shows how far each revision got. The dominant failure modes differ. \sys{} failed mostly at \emph{press down}: the handle turned but the latch did not release. It resolved this from the force recorded while pressing, where a sharp rise in resistance marks the end stop, and then fixed the remaining failures in pulling and releasing. The vision-only agent failed mostly at \emph{grasp} and \emph{pull open}. It adjusted the grasp and commanded press angle using images alone and could press the handle down, but the handle repeatedly slipped out of the gripper before the door opened. Table~\ref{tab:door} documents the failures and corresponding program revisions in both histories. The revision after execution~16 adds an angle condition for accepting an end stop and a force-drop condition before pulling. These checks change how contact evidence governs transitions between actions, illustrating refinement of the force-feedback logic itself.

%% file: figures/exp_adaptation_curve.tex
\begin{tikzpicture}
  \definecolor{cube}{HTML}{1B9E77}
  \definecolor{cyl}{HTML}{4A3AA7}
  \definecolor{ink}{HTML}{404040}
  \begin{axis}[
      width=\linewidth, height=4.0cm,
      font=\sffamily\scriptsize,
      xmin=-0.3, xmax=4.3, ymin=-22, ymax=400,
      xtick={0,1,2,3,4}, xticklabels={Start,1,2,3,4},
      ytick={0,100,200,300},
      xlabel={Attempt}, ylabel={Distance to target (mm)},
      xlabel style={yshift=3pt}, ylabel style={yshift=-4pt},
      axis lines=left, axis line style={ink, line width=0.5pt, -}, tick style={ink, line width=0.5pt},
      tick align=outside, major tick length=2pt,
      ymajorgrids, grid style={black!10, line width=0.4pt},
      every axis plot/.append style={line width=0.9pt},
      mark size=1.5pt,
      legend style={font=\sffamily\scriptsize, draw=none, fill=white, fill opacity=0.85, text opacity=1,
                    at={(1,1)}, anchor=north east, inner sep=1pt, column sep=3pt, row sep=-1pt},
      legend columns=2, legend cell align=left,
      legend image code/.code={\draw[mark repeat=2, mark phase=2] plot coordinates {(0cm,0cm) (0.25cm,0cm) (0.5cm,0cm)};},
    ]
    \fill[black!7] (axis cs:-0.3,-22) rectangle (axis cs:4.3,0);
    \addplot[cube, densely dotted, forget plot] coordinates {(0,377) (1,6)};
    \addplot[cube, densely dotted, forget plot] coordinates {(0,377) (1,136)};
    \addplot[cyl,  densely dotted, forget plot] coordinates {(0,349) (1,204)};
    \addplot[cyl,  densely dotted, forget plot] coordinates {(0,364) (1,59)};
    \addplot[cube, only marks, mark=square,  mark options={fill=white}, forget plot] coordinates {(0,377)};
    \addplot[cyl,  only marks, mark=*,       mark options={fill=white}, forget plot] coordinates {(0,349)};
    \addplot[cyl,  only marks, mark=square,  mark options={fill=white}, forget plot] coordinates {(0,364)};
    \addplot[cube, solid,  mark=*, forget plot]                                    coordinates {(1,6) (2,0)};
    \addplot[cube, dashed, mark=square, mark options={solid, fill=white}, forget plot] coordinates {(1,136) (2,105) (3,0)};
    \addplot[cyl,  solid,  mark=*, forget plot]                                    coordinates {(1,204) (2,19) (3,0.6) (4,0)};
    \addplot[cyl,  dashed, mark=square, mark options={solid, fill=white}, forget plot] coordinates {(1,59) (2,0)};
    \addlegendimage{cube, solid, line width=1.6pt}           \addlegendentry{Cube}
    \addlegendimage{ink, solid, mark=*}                      \addlegendentry{Whiteboard}
    \addlegendimage{cyl, solid, line width=1.6pt}            \addlegendentry{Cylinder}
    \addlegendimage{ink, dashed, mark=square, mark options={solid, fill=white}} \addlegendentry{Woodboard}
  \end{axis}
\end{tikzpicture}

%% file: figures/exp_door_iterations.tex
\begin{tikzpicture}[x=1pt, y=1pt, font=\sffamily\scriptsize]
  \definecolor{forca}{HTML}{2A78D6}   %
  \definecolor{vision}{HTML}{EB6834}
  \definecolor{muted}{HTML}{5A5A5A}
  \definecolor{pg0}{HTML}{E8A9A1}\definecolor{pg1}{HTML}{EDC09B}\definecolor{pg2}{HTML}{F0D69D}
  \definecolor{pg3}{HTML}{DEDDA0}\definecolor{pg4}{HTML}{C3DAA2}\definecolor{pg5}{HTML}{A3CF9F}
  \pgfdeclareverticalshading{doorprogressv}{100bp}{color(0bp)=(pg5); color(29.17bp)=(pg5);
    color(37.5bp)=(pg4); color(45.83bp)=(pg3); color(54.17bp)=(pg2); color(62.5bp)=(pg1);
    color(70.83bp)=(pg0); color(100bp)=(pg0)}
  \pgfmathsetmacro\xc{\linewidth/2}           %
  \def\bandw{54}                              %
  \def\rowh{12.5}                             %
  \def\pitch{10.5}\def\gap{4}                 %
  \tikzset{
    mk/.style={circle, minimum size=9pt, inner sep=0pt, text=white,
               font=\sffamily\bfseries\fontsize{5.5}{5.5}\selectfont},
    head/.style={inner sep=1pt, font=\sffamily\scriptsize\bfseries, text height=1.6ex, text depth=0.45ex},
  }
  \def\doorstage#1#2#3#4{%
    \def\dN{0}\foreach \dI in {#4}{\xdef\dN{\the\numexpr\dN+1\relax}}%
    \foreach \dI [count=\dK from 0] in {#4}{%
      \pgfmathsetmacro\dX{#3 < 0 ? \xc-\bandw/2-\gap-4.5-(\dN-1-\dK)*\pitch : \xc+\bandw/2+\gap+4.5+\dK*\pitch}%
      \node[mk, fill=#1, name=#1-\dI] at (\dX, {-(#2+0.5)*\rowh}) {\dI};
    }}

  \shade[shading=doorprogressv, rounded corners=2pt] (\xc-\bandw/2, 0) rectangle (\xc+\bandw/2, -6*\rowh);
  \foreach \lab [count=\s from 0] in {Approach, Grasp, Press down, Pull open, Release, \textbf{Success}}
    \node[inner sep=0pt, text=black, text height=1.6ex, text depth=0.45ex] at (\xc, {-(\s+0.5)*\rowh}) {\lab};

  \node[head, anchor=south]                at (\xc, 1.5)                   {Task progress};
  \node[head, anchor=south east, text=forca]  at (\xc-\bandw/2-\gap, 1.5)  {FORCA};
  \node[head, anchor=south west, text=vision] at (\xc+\bandw/2+\gap, 1.5)  {Vision-only};

  \doorstage{forca}{0}{-1}{1,2,9}
  \doorstage{forca}{2}{-1}{3,4,5,6,7,8,10,16}
  \doorstage{forca}{3}{-1}{11,14}
  \doorstage{forca}{4}{-1}{12,13,15}
  \doorstage{forca}{5}{-1}{17,18,19,20}

  \doorstage{vision}{0}{1}{1,2}
  \doorstage{vision}{1}{1}{3,4,5,6,9,10}
  \doorstage{vision}{2}{1}{7,8,11,12}
  \doorstage{vision}{3}{1}{13,14,15,16,17,18,19,20}

  \node[mk, fill=forca] (leg1) at (\xc+\bandw/2+\gap+4.5, -5*\rowh) {\itshape n};
  \node[mk, fill=vision, anchor=west] (leg2) at ([xshift=1.5pt]leg1.east) {\itshape n};
  \node[anchor=west, inner sep=0pt, align=left, text=muted, font=\sffamily\scriptsize] at ([xshift=3pt]leg2.east)
    {Iteration\\(program version)};
\end{tikzpicture}

%% file: Sections/5-Conclusion.tex
\section{Limitations}
\label{sec:limitations}
Our evaluation leaves several open questions about the cost and scope of deployment-time adaptation.

(i)~\emph{Iteration cost.}
Each iteration takes several minutes, with agent reasoning accounting for most of the time (Table~\ref{tab:rollouts}). The current refinement process also relies on real-robot rollouts. Future work could investigate ways to reduce reasoning overhead and whether preliminary program refinement in simulation can reduce physical trials, while evaluating how well the resulting programs transfer to the robot.

(ii)~\emph{Contact sensing.}
\sys{} relies on wrist wrench measurements for contact feedback. These measurements capture the combined effects of contact but do not directly reveal where contact occurs or how forces are distributed across the fingers. This limits the contact information available to the agent. Future work could explore distributed tactile sensing and evaluate its usefulness for contact diagnosis and finer manipulation.

(iii)~\emph{Task complexity.}
For tasks involving longer action sequences or tighter force constraints, the agent may struggle to diagnose failures and revise the program effectively, failing to produce a successful program even after repeated attempts. Future work could investigate whether improved reasoning and tool use help the agent solve such tasks.

\section{Conclusion}
\label{sec:conclusion}
We presented \sys{}, a coding agent that learns at deployment time by generating and revising robot programs using force feedback. Our real-robot experiments show that \sys{} can refine geometric estimates through contact, adapt existing programs to changed physical conditions, and maintain contact during wiping and peeling, without task demonstrations or task-specific model training.

We view this as a step toward deployment-time learning in which physical experience accumulates in executable, inspectable programs. Over repeated deployments, an agent could retain useful contact routines together with evidence about when they work, then revise them as conditions change. Further progress requires more reliable reasoning from contact signals and a more efficient revision process, so that agents can solve complex tasks with fewer physical trials and manual resets. Policies learned through imitation or built on vision-language-action models could also serve as skills within these programs. The agent would coordinate sensing and skill execution, and revise the program based on execution feedback.